\documentclass[10pt, a4paper]{article}
\usepackage{ltc23}
\usepackage{graphicx}
\usepackage{multirow}
\usepackage{hyperref}
\usepackage{color,soul}
\usepackage{tcolorbox}

\providecommand{\keywords}[1]
{
  \small	
  \textbf{\textit{Keywords: }} #1
}
\title{Detection of depression on social networks using transformers and ensembles}

\name{Ilija Tavchioski$^{\ast}$, Marko Robnik-\v{S}ikonja$^{\ast}$, Senja Pollak$^{\dagger}$} 
\address{ $^{\ast}$ University of Ljubljana, Faculty of Computer and Information Sciences\\
                  Ve\v{c}na pot 113, 1000 Ljubljana, Slovenia\\
                  $^{\dagger}$ Jozef Stefan Institute \\ 
                  Jamova Cesta 39, 1000 Ljubljana, Slovenia\\
                  }
\abstract{As the impact of technology on our lives is increasing, we witness increased use of social media that became an essential tool not only for communication but also for sharing information with community about our thoughts and feelings. This can be observed also for people with mental health disorders such as depression where they use social media for expressing their thoughts and asking for help. This opens a possibility to automatically process social media posts and detect signs of depression. We build several large pre-trained language model based classifiers for depression detection from social media posts. Besides fine-tuning BERT, RoBERTA, BERTweet, and mentalBERT were also construct two types of ensembles. We analyze the performance of our models on two data sets of posts from social platforms Reddit and Twitter, and investigate also the performance of transfer learning across the two data sets. The results show that transformer ensembles improve over the single transformer-based classifiers.}

\begin{document}

\maketitleabstract

\begin{tcolorbox}[width=\linewidth, colback=white!95!black]
The paper was presented in the scope of the LTC2023 April 21-23, 2023, Poznań, Poland, and published in:
\textit{Proceedings of 10th Language \& Technology Conference: Human Language Technologies as a Challenge for Computer Science and Linguistics, pages 282-287.}
\end{tcolorbox}
\section{Introduction}
\label{ch1}
Depression is one of the most widespread mental health disorders. According to the World Health Organization, it has the second highest number of affected people after anxiety, with 284 million cases worldwide \cite{james2018global}. Depression is affecting people's behaviour, mood, and feelings, but also affects their productivity at work and their relationships. If left untreated the depression can lead to serious consequences such as suicide, which is the case for 800{,}000 people every year \cite{salas2022detecting}. The bright side is the fact that the depression is treatable and its early detection is beneficial in the success of the treatment.
\par
During the past decade, the use of social media rapidly grew, and social platforms digitized  our society and human interactions. In many cases, people use social media to  express their thoughts and feelings and also to share important moments of their lives.
%OLD:With the rise of technology in the past two decades, social media is close to become the true digitization of our society. Nowadays, social media is a more preferable tool for expressing our thoughts and feelings, sharing important moments in our lives and etc.
 Marriott and Buchanan \cite{MARRIOTT2014171} showed that the expression of a person's  personality online is very similar or even identical to its expression  offline, hence presenting a possibility to infer knowledge of people's personalities along with mental health-related issues.
\par
We apply natural language processing (NLP) methods to social media posts, aiming to detect signs of depression and improve an early stage depression detection of depression, which is highly beneficial for its successful treatment. We build a number of transformer based classifiers and their ensembles and analyse classifiers' performance on two depression data sets. Besides direct assessment of depression in each dataset we also test cross-data set transfer. 
\par
The paper is structured into six sections. We present the related work in Section \ref{ch2}. In Section \ref{ch3}, we describe the problem and datasets. The methodology is presented in Section \ref{ch4}, while the  experiments  results are described in Section \ref{ch5}. Conclusions and plans for future work are presented in Section \ref{ch6}.

\section{Related work}
\label{ch2}
Several NLP researchers have tried to detect depression and related mental issues from social media, foremost from Twitter but also Reddit. \cite{coppersmith-etal-2016-exploratory} tried to detect suicide intentions in tweets using logistic regression with character n-gram features. \cite{arabic} checked depression in both English and Arabic social media posts  by extracting sparse features and constructing vector representations. They tested several classification methods such as Random Forest, Naive Bayes, AdaBoost and linear SVM. Later deep learning methods, such as bidirectional LSTM neural networks with attention \cite{blstm+attention} produced better performance. For Reddit, \cite{trifan2020understanding} proposed a SVM classifier with stochastic gradient descent using TF-IDF weighted feature vectors. Recently, the transformer-based architectures become the primary deep learning method for text analysis. \cite{acl-primer1} used several transformer-based methods such as RoBERTa, ALBERT, and DistilBERT and achieved decent results on multi-class data sets where each class corresponds to the level of depression expressed in the post. Some researchers combined  transformer-based methods with a combination of TF-IDF weighted vector representations and knowledge-graph based embeddings \cite{tavchioski2022e8}. A similar shared task aims to classify users as early as possible based on their history on Reddit \cite{clef}; one of the proposed approaches \cite{tavchioski2022early} constructed document embeddings using sentence-BERT \cite{reimers-2019-sentence-bert} and used logistic regression for classification.

\section{Problem description and data sets}
\label{ch3}
In this section, we define the depression detection problem and and two datasets used in our experiments along with their the distributions of class values (see Table \ref{tab:data-acl} for Reddit and Table \ref{tab:data-twitter} for Twitter dataset).

\subsection{Problem description}
The problem is defined as follows. We are given a set of social media posts $D = \{d_1, d_2, ..., d_n\}$ along with their respective labels $L = \{l_1, l_2, ..., l_n\}$, where the labels correspond to the level of depression signs present in posts. The goal is to train a prediction model that will label new posts as accurately as possible.

\subsection{Data description}
We used two English datasets in experiments. The Reddit dataset \cite{depsignacl1} is composed of posts from the Reddit social platform, mostly from subreddits like “r/stress”, “r/loneliness”, but also from “r/Anxiety”, “r/depression” etc. The second dataset is composed mosty of posts from the social platform Twitter \cite{kaggle}. 
The Twitter dataset contains short posts, some unrelated to depression. The Reddit dataset contains longer longer expressing persons' feelings in a deeper way. %OLD TEXT: Both data sets are in English and while the Twitter data set is more general one, meaning the posts are shorter and can be totally unrelated to depression, the Reddit data set is more concrete, the posts are longer and are expressing the person's feelings in a more deeper way.
The Reddit posts were annotated by two domain specialists into three classes corresponding to the level of depression signs in the post.
\begin{itemize}
    \item \textbf{Level 0 (Not depressed)} - there are no signs of depression in the post; the statements in the post are either irrelevant concerning depression or are related to giving help or motivation to people with depression.
    \item \textbf{Level 1 (Moderate)} - a post contains moderate signs of depression. These posts are related to change of feelings, but they show signs of improvement and hope.
    \item \textbf{Level 2 (Severe)} - these posts contain severe signs of depression. They are often related to serious suicide thoughts, disorder conditions or past suicide attempts.
\end{itemize}
\begin{table}[h]
  \centering
  \label{tab:data-acl}
  \begin{tabular}{ l | ccc}
  \hline
  Data set & Training & Validation & Test\\

  \hline
   Level 0 & 1659 (22\%)  &312 (23\%) & 2306 (51\%)\\
%    \textit{Not depr.} & & & \\

 Level 1 & 5140 (68\%) &879 (66\%)  & 1830 (41\%)\\
% \textit{Moderate} & & & \\

  Level 2  &  758 (10\%) & 143 (11\%) & 360 (8\%) \\
 % \textit{Severe} & & & \\
  \hline
  Total & 7557 & 1334 & 4496 \\

\end{tabular}
\caption{Reddit depression detection label distribution.}
\end{table}
The Twitter data set contains four depression labels corresponding to different levels of depresion signs. \textit{Level 0}  corresponds to the \textit{Level 0} from the previous dataset, i.e. Not depressed. \textit{Level 2} and \textit{Level 3} corresponds to \textit{Level 1} and \textit{Level 2} of the Reddit dataset, respectively, expressing Moderate and Severe depression. The Twitter dataset \textit{Level 1} is similar to  \textit{Level 2} label, where statements reflect a slight change of person's feelings, but do not drastically affect the person's mood. We call this label \textit{Change of feelings}.

\begin{table}[h]
  \centering
  \label{tab:data-twitter}
  \begin{tabular}{ c | c  c  c }
  Data set & Training & Validation & Test \\
  \hline
  Level 0 & 19095 (41\%) & 4118 (41\%) & 3989 (40\%)\\
 %    \textit{Not depr.} & & & \\
  Level 1 & 2408 (5\%) &  507 (5\%) & 537 (5\%) \\
%   \textit{Ch. feel.} & & & \\
  Level 2  & 20069 (43\%)  & 4272 (42\%)  & 4328 (43\%)\\
%   \textit{Moderate} & & & \\
  Level 3 & 6931 (15\%) & 1498 (14\%) & 1539 (15\%) \\
%   \textit{Severe} & & & \\
  \hline 
  Total & 46167 & 10045  & 10016\\
  
\end{tabular}
\caption{Twitter depression detection label distribution.}
\end{table}

\section{Depression prediction models}
\label{ch4}
In this section, we present the constructed models, starting with baseline methods, and followed by transformer and ensemble models, We end the presentation with cross/domain transfer models.

\subsection{Baseline methods}
For comparison of our models' performances we apply three baseline models.
\begin{description}
\item \textbf{Majority classifier} returns 
the most frequent label in the training set.
\item \textbf{TF-IDF } method
uses the logistic regression classifier from \textit{scikit-learn} \cite{scikit-learn} library with the default parameters using TF-IDF weighted feature extracted from the posts.
\item \textbf{Doc2Vec}
We used doc2vec method \cite{doc2vec} for construction of document embeddings for the posts and applied the logistic regression classifier with default parameters.
\end{description}

\subsection{Transformer-based models}
We tested several models based on large pre-trained language models similar to BERT \cite{bert} with a \textit{softmax} layer on top of the final hidden vector corresponding to the \textit{CLS} token with \textit{L} nodes, where \textit{L} corresponds to the number of labels in the dataset. The last \textit{softmax} layer returns label probabilities and the label with the highest probability is returned as prediction. The hyper-parameters  used for fine-tuning were chosen based on the validation set. The loss function used was the cross-entropy. The details for each of the for BERT-like models are presented below.

\begin{description}
\item \textbf{BERT}
The BERT base model \cite{bert} was pre-trained on the English Wikipedia (2{,}500 million words) and the BookCorpus (800 million words) \cite{bookcorpus} on two tasks. In the  masked language modelling task 15\% of the words were masked and the goal was to predict them. The second pre-training task is the next sentence prediction where given two sentences the goal is to predict is the second is the successor of the first. The model is composed of 14 stacked encoder blocks with 12 self-attention heads and the the vector representation of 768 dimensions. The total number of parameters is 110 million. We used the model and its \textit{tokenizer} (an algorithm for converting the text into a sequence of tokens) from the \textit{HuggingFace} framework, named \textit{bert-base-cased}.
\item \textbf{RoBERTa}
 \cite{liu2019roberta} is a BERT-like model with 24 encoder blocks, with vector representation of 1024 and 16 self-attention heads, with a total of %OLD:ahieving a
355 million parameters. It is pre-trained on larger corpus; in addition to English Wikipedia and BookCorpus, the models was also pre-trained on data from CC-News, OpenWebText, and Stories. RoBERTa does not use the next sentence prediction task in pre-training and adjusts the first task with dynamic masking, where the masking pattern is different for each input sequence. The model was pre-trained for longer and with larger batch sizes. We used the  \textit{HuggingFace} implementation, named \textit{roberta-base}. 
\item \textbf{mentalBERT}
While BERT and RoBERTa are general models,  pre-trained on general texts, for depression detection we used the mentalBERT model \cite{mentalBERT} which may be better suitable, since it is additionally pre-trained on mental health related data from the Reddit social platform. The model has the same architecture as the BERT model and the same tasks for pre-training, but the authors adjusted the BERT model by continuing the pre-training process with the new data from Reddit. The pretraining data of this model does not overlap with our Reddit dataset. It was was collected from subreddits  “r/SuicideWatch”, “r/offmychest”,  “r/Anxiety”, “r/mentalhealth”, “r/bipolar”, and, “r/mentalillness/”,  around two years prior to the collection of our Reddit dataset.  The model was taken from  \textit{HuggingFace} where it is named \textit{mental/mental-bert-base-uncased}.
\item \textbf{BERTweet}
As we  also use a dataset from the Twitter social network, we also tested the BERTweet model \cite{BERTweet} which uses the same pre-training approach as RoBERTa and the same architecture as BERT. The main difference  is that the model is pre-trained solely on Twitter data composed 850 million tweets in English. We used the  \textit{HuggingFace} \footnote{\url{https://huggingface.co/}} models named \textit{vinai/bertweet-base}. 

\end{description}

\subsection{Transformer ensembles}
We combined several standalone BERT-based models from above and formed different ensembles. We experimented with four combinations of baseline BERT models  (G-general model, M-mentalBERT, T-BERTweet). As general models (pre-trained on general texts), i.e. RoBERTa and BERT, we selected the one with better performance which in most cases is RoBERTa. The combinations are named as follows.
\begin{itemize}
    \item \textbf{GMT} - RoBERTa/BERT, mentalBERT and BERTweet.
    \item \textbf{GT} - RoBERTa/BERT and BERTweet.
    \item \textbf{GM} - RoBERTa/BERT and mentalBERT.
    \item \textbf{MT} - mentalBERT and BERTweet.
\end{itemize}

We combined the ensemble member predictions in two ways.
\begin{description}
  \item \textbf{Averaging ensembles (AE)}
calculate the output as the average of the models' probabilities. 
\item \textbf{Bayesian ensembles (BE)}
\cite{bayesian-ensemble} applies Bayesian framework and outputs the label  with the highest probability.
\end{description}

\subsection{Cross-data set transfer}
We have datasets from two different platforms (Twitter and Reddit), with  similar purpose but different in several aspects. To test the knowledge transfer between the datasets, we first fine-tuned BERT models on one data set, and then additionally on the second one, using the final model for the prediction. To align the labels in the datasets we merged two classes in the Twitter data set (\textit{Label 1} and \textit{Label 2}), since they are the most similar and only $5 \%$ of the data is labeled with \textit{Label 1}.

\section{Experiments and results}
\label{ch5}
In this section, we first present the evaluation metrics and selection of hyper-parameters, followed by the results on the test sets. We used the validation set to select the best hyper-parameters. For each setting, we repeated the experiments three times and took the mean of the results.

\subsection{Evaluation metrics}
The following metrics were used:
\begin{itemize}
    \item \textbf{Precision} - represents the percentage of correctly predicted instances out of all positive predictions.
    \item \textbf{Recall} - represents the percentage of correctly predicted instances out of all instances that were labeled as positive.
    \item \textbf{F1-score} is a harmonic mean of the precision and recall. 
    \[F_1 = \frac{2 \times precision \times recall}{precision + recall}\]
    \item \textbf{Accuracy} - represents the classification accuracy of predictions.
\end{itemize}
Since both problems are multi-class classification problems, we used weighted average of the scores for each class according to its number of instances.

\subsection{Hyper-parameter selection}
We tuned the hyper-parameters of all models on the validation set. We considered the following hyper-parameters.
\begin{itemize}
    \item \textbf{Batch size (BS)} -- $BS \in \{8, 16, 32\}$
    \item \textbf{Learning Rate (LR)} -- $LR \in \{10^{-3}, 10^{-4}, 5 \cdot 10^{-5}, 10^{-5}\}$
    \item \textbf{Number of epochs (NE)} -- $NE \in \{5, 10, 15\}$
\end{itemize}
The obtained hyper-parameters are presented in Table \ref{tab:parameters}.
\begin{table}[h]
\centering

  \label{tab:parameters}
  \begin{tabular}{l  c  c  c  c}
  \hline
  Model & Dataset & BS & NE & LR \\
  \hline
  BERT & Reddit & 32 & 15 & $5 \cdot 10^{-5}$ \\
  
  RoBERTa & Reddit & 8 & 15 & $10^{-5} $\\
  
  mentalBERT & Reddit & 32 & 15 & $10^{-4} $\\
  
  BERTweet & Reddit & 16 & 15 & $5 \cdot 10^{-5}$ \\

  \hline
  BERT & Twitter & 32 & 15 & $5 \cdot 10^{-5} $\\
  
  RoBERTa & Twitter & 32 & 15 &$ 5 \cdot 10^{-5}$ \\
  
  mentalBERT & Twitter & 32 & 10 & $5 \cdot 10^{-5}$ \\
  
  BERTweet & Twitter & 16 & 5 & $5 \cdot 10^{-5}$ \\
  \hline
  
\end{tabular}
\caption{The values of hyper-parameters used for BERT models, based on the validation set. We tuned batch size (BS), learning rate (LR) and number of epoch (NE).}
\end{table}

Finally, we fine-tuned the models using both the training and validation data and the same hyper-parameters were used for the transfer learning and ensembles.

\subsection{Results}
For each setting, we conducted two types of experiments, one without transfer learning from the other dataset (the first line for  each model), and the other with transfer learning (the second line---in italics---for each model). Results are presented in  Tables (\ref{tab:results-acl} and \ref{tab:results-twitter}), for Reddit and Twitter, respectively.

\subsubsection{Reddit result}
\begin{table}[h]
  \centering
  \label{tab:results-acl}
  \begin{tabular}{|| c | c | c | c | c ||}
  \hline
   & Model & Acc & F1 & SD\\
  \hline
  \hline
  \multirow{3}{3em}{ Base-\\lines}&majority & 0.513&0.348&0.0000\\\cline{2-5}
  &doc2vec & 0.459&0.459&0.0051\\\cline{2-5}
  &TF-IDF & 0.519&0.576&0.0000\\\hline
   \hline
  \multirow{8}{3em}{Trans-\\formers}&\multirow{2}{*}{mentalBERT}&0.577&0.577&0.0050\\\cline{3-5}
  &&\textit{0.569}&\textit{0.565}&\textit{0.0079}\\\cline{2-5}\cline{2-5}
  &\multirow{2}{*}{RoBERTa}&0.557&0.563&0.0035\\\cline{3-5}
  &&\textit{0.532}&\textit{0.537}&\textit{0.0110}\\\cline{2-5}\cline{2-5}
  &\multirow{2}{*}{BERT}&0.559&0.561&0.0057\\\cline{3-5}
  &&\textit{0.564}&\textit{0.559}&\textit{0.0051}\\\cline{2-5}\cline{2-5}
  &\multirow{2}{*}{BERTweet}&0.560&0.561&0.0072\\\cline{3-5}
  &&\textit{0.557}&\textit{0.553}&\textit{0.0068}\\\hline
  \hline
  \multirow{8}{3em}{ AE}&\multirow{2}{*}{GMT}&\textbf{0.592}&\textbf{0.592}&\textbf{0.0024}\\\cline{3-5}
  &&\textit{0.586}&\textit{0.580}&\textit{0.0040}\\\cline{2-5}\cline{2-5}
  &\multirow{2}{*}{GM}&0.592&0.591&0.0031\\\cline{3-5}
  &&\textit{0.583}&\textit{0.575}&\textit{0.0057}\\\cline{2-5}\cline{2-5}
  &\multirow{2}{*}{MT}&0.584&0.579&0.0057\\\cline{3-5}
  &&\textit{0.579}&\textit{0.571}&\textit{0.0093}\\\cline{2-5}\cline{2-5}
  &\multirow{2}{*}{GT}&0.579&0.580&0.0030\\\cline{3-5}
  &&\textit{0.577}&\textit{0.569}&\textit{0.0021}\\\hline
  \hline
  \multirow{8}{3em}{ BE}&\multirow{2}{*}{GMT}&0.588&0.590&0.0037\\\cline{3-5}
  &&\textit{0.564}&\textit{0.567}&\textit{0.0063}\\\cline{2-5}\cline{2-5}
  &\multirow{2}{*}{GM}&0.545&0.556&0.0045\\\cline{3-5}
  &&\textit{0.496}&\textit{0.513}&\textit{0.0209}\\\cline{2-5}\cline{2-5}
  &\multirow{2}{*}{MT}&0.525&0.541&0.0125\\\cline{3-5}
  &&\textit{0.525}&\textit{0.537}&\textit{0.0051}\\\cline{2-5}\cline{2-5}
  &\multirow{2}{*}{GT}&0.568&0.571&0.0086\\\cline{3-5}
  &&\textit{0.565}&\textit{0.562}&\textit{0.0069}\\\hline

\end{tabular}
\caption{Results of baseline models, standalone transformers, averaging ensembles (AE) and Bayesian ensembles (BE) on the Reddit dataset. For each setting, the results with transfer learning from Twitter data set are in \textit{italics} and the results without transfer learning are without italics.}
\end{table}

In Table \ref{tab:results-acl}, we present  the results  on the Reddit dataset introduced in Section \ref{tab:data-acl}. Not surprisingly, the lowest performance methods were the baselines, the majority classifier, doc2vec and TF-IDF with logistic regression. The standalone transformers are much more successful, with mentalBERT (pretrained on Reddit data) being the best in this group, followed by RoBERTa, BERT and BERTweet. For majority of ensemble models, RoBERTa is used as the general model, except in the transfer learning experiments using averaging and Bayesian ensembles, where BERT performed better. For Bayesian ensembles, we can see that the GMT (RoBERTa, mentalBERT and BERTweet) and GT (RoBERTa and BERTweet) improved the results in comparison to the single BERT models, but the other combinations (MT (mentalBERT and BERTweet) and GM (RoBERTa and mentalBERT)) did not. Finally, the averaging ensembles were consistent in improving the results where the GMT (RoBERTa, mentalBERT and BERTweet) ensemble had the highest performance with F1-score of $0.592$ and standard deviation of $0.0024$. Example of the improvement can be the following Reddit post "\textit{My dad had to explain to me that high school parties are a real thing. : I graduated 2 years ago and I genuinely thought high school parties were only from TV because I was never invited to or told about one. I'm not going to college either so to my knowledge, college parties don't exist either. I was going to commuter college but dropped out. I can't afford to go back for the forseeable future either. Social isolation is really doin it to me.}", where the RoBERTa model assessed the post with moderate level of depression while the ensemble GMT classified the post with the Level 0 which is the correct label. Regarding the transfer learning from the Twitter dataset, we can see that it did not lead to improved results.
In general, the standard deviations are small, and the ensembles mostly further reduce the variance compared to standalone models. 

\par
The Reddit data set was part of the shared task \cite{depsignacl} organized by ACL and the best scores achieved in the competitions varied from $0.60$ to $0.64$ which testifies of decent performance our models exhibit. Note that the presented results here are based on the performance from the development set instead of the test set due to the missing gold labels of the unavailable official test set. The development and the test set were composed in a similar manner. Although many of the methods that were used are transformer-based, our ensemble methods introduce new combinations of transformer-based models such as combining mentalBERT and BERTweet. The code for the experiments can be find at \url{https://gitlab.com/teletton/diploma}.

\subsubsection{Twitter results}
\begin{table}[t!]
  \centering
  \label{tab:results-twitter}
  \begin{tabular}{|| c | c | c | c | c ||}
  \hline
   & Model & Acc & F1  & SD\\
  \hline
  \hline
  \multirow{3}{3em}{  Base-\\lines}&majority & 0.416&0.245&0.0000\\\cline{2-5}
  &doc2vec &0.683&0.680&0.0009\\\cline{2-5}
  &TF-IDF &0.730&0.728&0.0000\\\hline
  \hline
  \multirow{8}{3em}{Trans-\\formers}&\multirow{2}{*}{mentalBERT}&0.831&0.831&0.0023\\\cline{3-5}
  &&\textit{0.848}&\textit{0.848}&\textit{0.0004}\\\cline{2-5}\cline{2-5}
  &\multirow{2}{*}{RoBERTa}&0.852&0.852&0.0009\\\cline{3-5}
  &&\textit{0.865}&\textit{0.866}&\textit{0.0020}\\\cline{2-5}\cline{2-5}
  &\multirow{2}{*}{BERT}&0.831&0.831&0.0008\\\cline{3-5}
  &&\textit{0.846}&\textit{0.846}&\textit{0.0003}\\\cline{2-5}\cline{2-5}
  &\multirow{2}{*}{BERTweet}&0.849&0.849&0.0008\\\cline{3-5}
  &&\textit{0.860}&\textit{0.860}&\textit{0.0043}\\\hline
  \hline
  \multirow{8}{3em}{ AE}&\multirow{2}{*}{GMT}&0.858&0.859&0.0018\\\cline{3-5}
  &&\textit{0.871}&\textit{0.871}&\textit{0.0005}\\\cline{2-5}\cline{2-5}
  &\multirow{2}{*}{GM}&0.851&0.851&0.0017\\\cline{3-5}
  &&\textit{0.857}&\textit{0.857}&\textit{0.0024}\\\cline{2-5}\cline{2-5}
  &\multirow{2}{*}{MT}&0.839&0.839&0.0029\\\cline{3-5}
  &&\textit{0.853}&\textit{0.853}&\textit{0.0013}\\\cline{2-5}\cline{2-5}
  &\multirow{2}{*}{GT}&0.858&0.858&0.0014\\\cline{3-5}
  &&\textit{\textbf{0.873}}&\textit{\textbf{0.873}}&\textit{\textbf{0.0005}}\\\hline
  \hline
  \multirow{8}{3em}{ BE}&\multirow{2}{*}{GMT}&0.849&0.849&0.0064\\\cline{3-5}
  &&\textit{0.857}&\textit{0.858}&\textit{0.0025}\\\cline{2-5}\cline{2-5}
  &\multirow{2}{*}{GM}&0.853&0.854&0.0003\\\cline{3-5}
  &&\textit{0.866}&\textit{0.866}&\textit{0.0016}\\\cline{2-5}\cline{2-5}
  &\multirow{2}{*}{MT}&0.835&0.835&0.0054\\\cline{3-5}
  &&\textit{0.849}&\textit{0.849}&\textit{0.0005}\\\cline{2-5}\cline{2-5}
  &\multirow{2}{*}{GT}&0.844&0.844&0.0020\\\cline{3-5}
  &&\textit{0.854}&\textit{0.854}&\textit{0.0017}\\\hline
  
\end{tabular}
\caption{Results of baseline models, simple transformers, averaging ensembles (AE) and Bayesian ensembles (BE) for the Twitter data set. For each setting, the results with transfer learning from Twitter data set are in \textit{italics} and the results without transfer learning are without italics.}
\end{table}

Table \ref{tab:results-twitter} presents the results on the Twitter dataset. Similarly to Twitter, the baseline methods lag behind standalone transformers, and ensembles. In standalone BERT-like models, RoBERTa  considerably outperformed other models, followed by BERTweet (pretrained on Twitter data), BERT and mentalBERT. The Bayesian ensembles show lower performance in comparison to the standalone BERT methods. As was the case in the Reddit dataset, the averaging ensembles  showed the best  performance with the GMT ensemble (RoBERTa, mentalBERT and BERTweet) giving the weighted F1-score of $0.859$. Again in comparison with the RoBERTa for example "texas at night is creepy", the RoBERTa model asigned the Level 1 class while the GMT model assigned the Level 2 class which is the correct one. In contrast to the previous dataset, the transfer learning did improve the results for almost all methods and the highest performing model was the GT (RoBERTa and BERTweet) averaging ensemble with transfer learning, obtaining the  weighted F1-score of $0.873$.

\section{Conclusion and future work}
\label{ch6}
The results show that fine-tuning large pre-trained language models can be successfully used to predict depression levels from social media. They outperform standard baselines such as majority classifier and logistic regression with doc2vec and TF-IDF weighted features. The use of models pre-trained on specific domain data turned out to be  better compared to transfer from other domain for standalone BERT models. The mentalBERT model and BERTweet model were successful for datasets from the domain where they were pretrained, showing benefit of such domain specific pretraining.  
\par
The Bayesian ensembles did not surpass much simpler averaging ensembles and even, in some cases, produced lower results than standalone BERT models. This may be due to the fact that all members of ensembles were relatively similar and Bayesian ensembles did not get a chance to learn and exploit their differences. 
%are more useful when the models make different mistakes\hl{I removed most of this paragraph, see comment, maybe Marko RS can suggest a better formulated text or we leave it out:}, which seems not to be our case.%OLD TEXT: this is not clear, i removed it and are relatively not very accurate which is not the case in our task since all models which are part of the ensembles are relatively high performance models for sentiment analysis.
\par
We also note the difference in the performance of transfer learning in the results of the two data sets, where  transfer learning improves the results on the Twitter data set but not on the Reddit dataset. One possible explanation may be that the Twitter data set is much larger and more general, which can have a strong impact on fine-tuning with the Reddit data set. On the other hand, the Reddit data set is much smaller with longer posts and can enrich the model with knowledge about different types of expressing feelings which can occur in tweets. The second explanation may be that due to merging of the labels, the models had easier task to determining the labels since there is one less class. % and there are not the two classes which were difficult to distinguish anymore.
\par
Our methods could be further improved in several ways. First, the input size is limited and  the model truncates longer input posts  which can lead to loss of information. In future work, we will consider the models with larger inputs such as Longformer \cite{longformer}. Also, we could try to enriching our set of ensemble  models with other BERT-based models.
%but also with ones that are not BERT-based, for example models that are not highly accurate but make different mistakes than the BERT does, which can lead to improvement in the Bayesian ensembles.
%\par
%In short summary, we showed that by using natural language processing we can detect level of depression signs in social media posts.

\section{Acknowledgments}
The research was supported by the Slovene Research Agency through research core funding no. P6-0411 and P2-103, as well as project no. J6-2581. 

%\nocite{*}

\bibliographystyle{ltc23}
\bibliography{xample23} 

\end{document}